\documentclass[letterpaper,11pt]{llncs} 

\usepackage{mathptmx}
\usepackage[scaled=.90]{helvet}
\usepackage[unicode]{hyperref}
\usepackage{mathtools,amssymb,amsfonts}
\usepackage{tikz}
\usepackage{array}
\usepackage{pifont}
\usepackage{engord}
\usepackage{fancyhdr}
\usepackage{fancyvrb}
\usepackage{pgfplots}
\usepackage{algorithm}
\usepackage{algpseudocode}
\usepackage[normalem]{ulem}
\usepackage[plain]{fancyref}
\usepackage[unicode]{hyperref}
\usepackage{graphicx,tabularx}
\usepackage{exscale}
\usepackage{relsize}
\usepackage{mathrsfs}
\usepackage{textcomp}
\usepackage{paralist}

\usepackage{subfigure}
\usepackage{multirow}

\usepackage{setspace}

\let\subparagraph\relax
\usepackage[compact]{titlesec}

\usetikzlibrary{calc}
\usetikzlibrary{shapes}
\usetikzlibrary{patterns}
\usetikzlibrary{positioning}
\usetikzlibrary{automata}
\usetikzlibrary{fit}
\usetikzlibrary{trees}
\usetikzlibrary{backgrounds}
\tikzset{
    position/.style args={#1:#2 from #3}{
        at=(#3.#1), anchor=#1+180, shift=(#1:#2)
    }
}

\newcommand*{\fancyrefdeflabelprefix}{def}
\newcommand*{\fancyrefthmlabelprefix}{thm}
\newcommand*{\fancyreflemlabelprefix}{lem}
\newcommand*{\fancyrefcorlabelprefix}{cor}
\newcommand*{\fancyrefalglabelprefix}{alg}
\newcommand*{\fancyreflnlabelprefix}{ln}
\frefformat{plain}{\fancyreflnlabelprefix}{line #1}
\Frefformat{plain}{\fancyreflnlabelprefix}{Line #1}
\frefformat{plain}{\fancyrefalglabelprefix}{Alg. #1}
\Frefformat{plain}{\fancyrefalglabelprefix}{Alg. #1}
\frefformat{plain}{\fancyrefdeflabelprefix}{definition #1}
\Frefformat{plain}{\fancyrefdeflabelprefix}{Definition #1}
\frefformat{plain}{\fancyrefthmlabelprefix}{theorem #1}
\Frefformat{plain}{\fancyrefthmlabelprefix}{Theorem #1}
\frefformat{plain}{\fancyreflemlabelprefix}{lemma #1}
\Frefformat{plain}{\fancyreflemlabelprefix}{Lemma #1}
\frefformat{plain}{\fancyrefcorlabelprefix}{corollary #1}
\Frefformat{plain}{\fancyrefcorlabelprefix}{Corollary #1}

\algnewcommand{\Break}{\textbf{break}}
\algnewcommand{\Continue}{\textbf{continue}}
\algnewcommand{\True}{\text{True}}
\algnewcommand{\False}{\text{False}}
\algnewcommand{\Not}{\textbf{not}~}
\algnewcommand\algorithmicto{\textbf{to}}
\algrenewtext{ForTo}[3]{\algorithmicfor\ #1 \gets #2 \algorithmicto\ #3 \algorithmicdo}

\floatname{}{Alg.}

\allowdisplaybreaks[4]

\providecommand{\n}[1]{{\color{black}#1}}

\expandafter\def\expandafter\normalsize\expandafter{%
    \normalsize
    \setlength\abovedisplayskip{5pt}
    \setlength\belowdisplayskip{5pt}
    \setlength\abovedisplayshortskip{0pt}
    \setlength\belowdisplayshortskip{0pt}
}

\begin{document}

\title{Computational Protein Design Using \\ AND/OR Branch-and-Bound Search\vspace{-10pt}}
\author{
  Yichao Zhou\inst{1} \and
  Yuexin Wu\inst{1} \and
  Jianyang Zeng\inst{1,2,}\thanks{
    Corresponding author: Jianyang Zeng.  Email:
    \href{mailto:zengjy321@tsinghua.edu.cn}{zengjy321@tsinghua.edu.cn}.
  }
}
\institute{
  Institute for Interdisciplinary Information Sciences, Tsinghua University, P.R.C.
  \and
  MOE Key Laboratory of Bioinformatics, Tsinghua University, P.R.C.
}
\maketitle
\begin{abstract}
  The computation of the global minimum energy conformation (GMEC) is an
  important and challenging topic in structure-based computational protein
  design.  In this paper, we propose a new protein design algorithm based on
  the AND/OR branch-and-bound (AOBB) search, which is a variant of the
  traditional branch-and-bound search algorithm, to solve this combinatorial
  optimization problem.  By integrating with a powerful heuristic function,
  AOBB is able to fully exploit the graph structure of the underlying residue
  interaction network of a backbone template to significantly accelerate the
  design process.  Tests on real protein data show that our new protein design
  algorithm is able to solve many problems that were previously unsolvable by
  the traditional exact search algorithms,  and for the problems that can be
  solved with traditional provable algorithms, our new method can provide a
  large speedup by several orders of magnitude while still guaranteeing to find
  the global minimum energy conformation (GMEC) solution.

  \textbf{Keywords:} protein design, AND/OR branch-and-bound, global minimum
  energy conformation, residue interaction network, mini-bucket heuristic
\end{abstract}

\section{Introduction} 
In a \emph{structure-based computational protein design} (SCPD) problem, we
aim to find a new amino acid sequence that accommodates certain structural
requirements and thus can perform desired functions by replacing several
residues from a wild-type protein template.  The SCPD has exhibited promising
applications in numerous biological engineering situations, such as enzyme
synthesis \cite{chen2009computational}, drug resistance prediction
\cite{frey2010predicting}, drug design \cite{gorczynski2007allosteric}, and
design of protein-protein interactions \cite{roberts2012computational}.

The aim of SCPD is to find the \emph{global minimum energy conformation}
(GMEC), that is, the global optimal solution of an amino acid sequence that
minimizes a defined energy function. In practice, the rigid body assumption
which anchors the backbone template is usually applied to reduce computational
complexity.  In addition, possible side-chain assignments for a residue are
further discretized into several known conformations, called the \emph{rotamer
library}. It has been proved that SCPD is NP-hard \cite{pierce2002protein} even
with the two aforementioned prerequisites. A number of heuristic methods have
been proposed to approximate the GMEC
\cite{street1999computational,kuhlman2000native,marvin2001conversion}.
Unfortunately, these heuristic methods can be trapped into local minima and may
lead to poor quality of the final solution. On the other hand, several exact
and provable search algorithms which guarantee to find the GMEC solution have
been proposed, such as Dead-End Elimination (DEE) \cite{desmet1992dead}, A*
search
\cite{leach1998exploring,lippow2007progress,donald2011algorithms,zhou2015massively},
tree decomposition \cite{xu2006fast}, branch-and-bound (BnB) search
\cite{hong2006protein,traore2013new,david2014computational}, and BnB-based
linear integer programming
\cite{althaus2002combinatorial,kingsford2005solving}.

In our protein design scheme, a set of DEE criteria
\cite{goldstein1994efficient,gainza2012protein} is first applied to prune the
infeasible rotamers that are provably not part of the GMEC solution.  \n{After
that, the AND/OR branch-and-bound (AOBB) search \cite{marinescu2009and}} \n{is
used to traverse} over the remaining conformational space and find the GMEC
solution.
In addition, we propose an elegant extension of this AND/OR branch-and-bound
algorithm to compute the top $k$ solutions within a user-defined energy cutoff
from the GMEC.  Our tests on real protein data show that our new protein design
algorithm can address many design problems which cannot be solved exactly
before, and for the problems that were solvable formerly, our new method can
achieve a significant speedup by several orders of magnitude.

\subsection{Related Work} 
The A* algorithm \cite{leach1998exploring,keedy2013osprey}
uses a priority
queue to store all the expanded states, which unfortunately may exceed the
hardware memory limitation \n{for large problems}.  AOBB, on the contrary, uses
depth-first-search strategy that only requires linear space complexity \n{with}
respect to the number of mutable residues.

The traditional BnB search algorithm \cite{hong2006protein} usually ignores the
underlying topological information of the residue interaction network
constructed based on the backbone template, while AOBB is designed
to exploit this property.

Although the tree decomposition method \cite{xu2006fast} utilizes the residue
interaction network,
the table allocated by its
dynamic programming routine may be too large to fit in memory.  To fix this
problem, AOBB adopts \n{the mini-bucket heuristic}
to prune \n{a large number of} states to speeds up the search process.

\section{Methods} 

\subsection{Overview} \label{sec:overview} 
Under the assumptions of rigid backbone structures and discrete side-chain
conformations, the structure-based computational protein design (SCPD) can be
formulated as a combinatorial optimization problem which aims to find the best
rotamer sequence $r = (r_1, \dots, r_n)$ that minimizes following objective
function:
\begin{equation}
  E_T(r) = E_0 +
    \sum_{i=1}^nE_1(r_i) +
    \sum_{i=1}^n\sum_{j=i+1}^n E_2(r_i, r_j)\,,
  \label{eq:obj}
\end{equation}
where $n$ stands for the number of mutable residues, $E_T(r)$ represents the
total energy of the system in which the rotamer assignment of the mutable
residues is $r$, \n{$E_0$ represents the template energy (i.e., the sum of the
backbone energy and the energy among non-mutable residues), $E_1(r_i)$
represents the self energy of rotamer $r_i$ (i.e., the sum of intra-residue
energy and the energy between $r_i$ and non-mutable residues), and $E_2(r_i,
r_j)$ is the pairwise energy between rotamers $r_i$ and $r_j$.}


\subsection{AND/OR Branch-and-Bound Search} \label{sec:aobb} 
\subsubsection{Branch-and-Bound Search} \label{sec:bnb} 

Suppose we try to find the global minimum value of the energy
function $E(r)$, in which $r \in R$ and $R$ is the conformational space of the
rotamers.  The BnB algorithm executes two steps recursively.  The first step is
called \emph{branching}, in which we split the conformational space $R$ into
two or more smaller spaces, i.e., $R_1$, $R_2$, \dots, $R_m$, where $R_1 \cup
R_2 \cup \dots \cup R_m = R$.  If we are able to find $\hat{r_i} = \arg\min_{r
\in R_i} E(r)$ for all $i \in \{1, 2, \dots, m\}$, we can compute the minimum
energy conformation in the conformational space $R$ by identifying one of
$\hat{r_i}$ that has the lowest energy.

The second step of BnB is called \emph{bounding}.  Suppose the current lowest
energy conformation is $\overline{r_i}$.  For any sub-space $R_j$, if we can
ensure that the lower bound of the energy of all conformations in $R_j$ is
greater than $E(\overline{r_i})$,  we can prune the whole sub-space $R_j$
safely.  The lower bound of the energy of the conformations in a given space
usually can be computed based on some heuristic functions.  \n{The BnB
algorithm performs the branching and bounding steps recursively until the
current conformational space contains only one single conformation.  A more
detailed introduction to branch-and-bound search can be found in Appendix
Section A1 \cite{zhou2015appendix}.}

\subsubsection{Residue Interaction Network} \label{sec:network} 
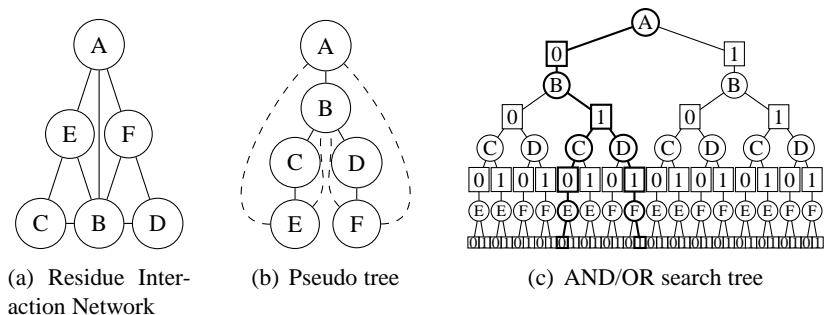
\begin{figure} 
  \centering
  \subfigure[Residue Interaction Network]{
    \begin{tikzpicture}[
        font=\small,
        node distance=1.0cm,
        node/.style={
          circle,
          draw,
        },
      ]
      \node[node] (A) {A};
      \node[node, below=1.7 of A] (B) {B};
      \node[node, left =0.1 of B] (C) {C};
      \node[node, right=0.1 of B] (D) {D};
      \node[node] (E) at ($(A) ! 0.5 ! (C)$) {E};
      \node[node] (F) at ($(A) ! 0.5 ! (D)$) {F};
      \path
        (A) edge (E) edge (F) edge (B)
        (B) edge (E) edge (F) edge (C) edge (D)
        (C) edge (E) edge (B)
        (D) edge (F) edge (B)
        ;
    \end{tikzpicture}
    \label{fig:network}
  }
  \subfigure[Pseudo tree]{
    \centering
    \begin{tikzpicture}[
        font=\small,
        node/.style={
          circle,
          draw,
        },
      ]
      \node[node] (A) {A};
      \node[node, below=0.15 of A] (B) {B};
      \node[node, below left =0.35 of B, xshift=9pt] (C) {C};
      \node[node, below right=0.35 of B, xshift=-9pt] (D) {D};
      \node[node, below=0.15 of C] (E) {E};
      \node[node, below=0.15 of D] (F) {F};
      \path
        (A) edge (B)
        (B) edge (C) edge (D)
        (C) edge (E)
        (D) edge (F)
        (E) edge[dashed, out=45,in=260] (B)
            edge[dashed, out=180, in=225](A)
        (F) edge[dashed, out=135,in=280] (B)
            edge[dashed, out=0, in=315](A)
        ;
    \end{tikzpicture}
    \label{fig:psedutree}
  }
  \subfigure[AND/OR search tree]{
    \centering
    \begin{tikzpicture}[
        cir/.style={
          circle,
          inner sep=0.8pt,
        },
        box/.style={
          rectangle,
          inner sep=1.8pt,
        },
        hi/.style={
          thick,
          shorten >=-0.6pt,
          shorten <=-0.6pt,
        },
        no/.style={
          thin,
          shorten >=-0.2pt,
          shorten <=-0.2pt,
        },
        font=\scriptsize,
        level distance=11.9pt,
        shorten >=-0.2pt,
        shorten <=-0.2pt,
        level 1/.style={sibling distance=67.2pt,box},
        level 2/.style={sibling distance=67.2pt,cir},
        level 3/.style={sibling distance=33.6pt,box},
        level 4/.style={sibling distance=16.8pt,cir},
        level 5/.style={sibling distance=8.4pt,box},
        level 6/.style={sibling distance=4.2pt,cir,font=\tiny,},
        level 7/.style={sibling distance=4.2pt,box, inner sep=0, minimum
        size=4.2pt,font=\tiny,},
      ]
      \tikzstyle{every node}=[draw]
        \node[cir,hi] (Root) {A}
          child[hi] { node {0}
            child { node {B}
              child[no] { node {0}
                child { node {C}
                  child { node {0}
                    child { node {E}
                      child { node {0} }
                      child { node {1} }
                    }
                  }
                  child { node {1}
                    child { node {E}
                      child { node {0} }
                      child { node {1} }
                    }
                  }
                }
                child { node {D}
                  child { node {0}
                    child { node {F}
                      child { node {0} }
                      child { node {1} }
                    }
                  }
                  child { node {1}
                    child { node {F}
                      child { node {0} }
                      child { node {1} }
                    }
                  }
                }
              }
              child { node {1}
                child { node {C}
                  child { node {0}
                    child { node {E}
                      child { node {0} }
                      child[no] { node {1} }
                    }
                  }
                  child[no] { node {1}
                    child { node {E}
                      child { node {0} }
                      child { node {1} }
                    }
                  }
                }
                child { node {D}
                  child[no] { node {0}
                    child { node {F}
                      child { node {0} }
                      child { node {1} }
                    }
                  }
                  child { node {1}
                    child { node {F}
                      child[no] { node {0} }
                      child { node {1} }
                    }
                  }
                }
              }
            }
          }
          child { node{1}
            child { node {B}
              child { node {0}
                child { node {C}
                  child { node {0}
                    child { node {E}
                      child { node {0} }
                      child { node {1} }
                    }
                  }
                  child { node {1}
                    child { node {E}
                      child { node {0} }
                      child { node {1} }
                    }
                  }
                }
                child { node {D}
                  child { node {0}
                    child { node {F}
                      child { node {0} }
                      child { node {1} }
                    }
                  }
                  child { node {1}
                    child { node {F}
                      child { node {0} }
                      child { node {1} }
                    }
                  }
                }
              }
              child { node {1}
                child { node {C}
                  child { node {0}
                    child { node {E}
                      child { node {0} }
                      child { node {1} }
                    }
                  }
                  child { node {1}
                    child { node {E}
                      child { node {0} }
                      child { node {1} }
                    }
                  }
                }
                child { node {D}
                  child { node {0}
                    child { node {F}
                      child { node {0} }
                      child { node {1} }
                    }
                  }
                  child { node {1}
                    child { node {F}
                      child { node {0} }
                      child { node {1} }
                    }
                  }
                }
              }
            }
          };
    \end{tikzpicture}
    \label{fig:aotree}
  }
  \caption{
    An example of constructing an AND/OR search tree.  (a) An example of a
    residue interaction network.  (b) The corresponding pseudo-tree of the
    residue interaction network in (a), in which dashed lines are non-tree
    edges.  (c) The full AND/OR search tree constructed from the pseudo-tree in
    (b), in which circle nodes represent OR nodes and rectangle nodes represent
    AND nodes.  An example of a solution tree for the AND/OR search tree in (c)
    is marked in bold.
  }
  \label{fig:aobb}
\end{figure}

\n{Traditional BnB algorithm can hardly exploit the underlying graph structure
of the residue-residue interactions}.
In a \n{real design problem},
some mutable residues can be relatively distant and thus the pairwise energy
terms in \Fref{eq:obj} between these residues are usually negligible.  Based on
this observation, we can construct a \emph{residue interaction network}, in
which each node represents a residue, and two nodes are connected by an
undirected edge if and only if the distance between them is less than a
threshold.  \Fref{fig:network} gives an example of such a residue interaction
network.

Consider a residue interaction network which contains two connected components
(i.e., two clusters of mutable residues at two distant positions).  Suppose
each residue has at most $p$ rotamers and the size of each connected component
is $q$.  Then the traditional BnB search needs to visit $O(p^{2q})$ nodes in
the worst case.  However, from the residue interaction network, we know that
two connected components are independent, which means that altering the
rotamers in one connected component does not affect the pairwise energy terms in
the other connected component.  So we can run the BnB search for each connected
component independently and then put the resulting minimum energy conformations
together to form the GMEC solution, which only needs to visit $O(p^q)$ nodes in
the worst case.

The independence requirement of connected components in a residue interaction
network is too strict in practice.  In fact, we can partition the whole network
into several independent connected components after choosing particular
rotamers in some residues.  For example, after fixing the rotamers for residues
$A$ and $B$ in the example shown in \Fref{fig:network}, we can obtain two
independent components $CE$ and $DF$.  Then we can use the aforementioned
method to reduce the size of search space and then search it using
branch-and-bound algorithm.  This is the major motivation of AND/OR
branch-and-bound (AOBB) search \cite{marinescu2009and}.

\subsubsection{AND/OR Search Space} \label{sec:ao} 

A \emph{pseudo-tree} \cite{freuder1985taking} of a graph (network) is a rooted
spanning tree on that graph in which every non-tree edge in the graph is
connected from a node to its offspring in the spanning tree.  In other words,
non-tree edges are not allowed to connect two nodes that are located in
different branches of the spanning tree.  \Fref{fig:psedutree} shows an example
of a pseudo-tree constructed based on the residue interaction network in
\Fref{fig:network}.

The pseudo-tree is a useful representation because for any node $x$ in the
tree, once all the side-chains of $x$ and its ancestors are fixed, all the
sub-trees rooted at the children of node $x$ are independent.  In other words,
altering the rotamers for the sub-tree rooted at a child of $x$ does not affect
the total energy of the another sub-tree.  Thus, the size of the search space
for all sub-trees rooted at children of node $x$ is proportional to the sum of
the sizes of these sub-tress rather than the product of their sizes as in the
traditional BnB algorithm.  Therefore, AOBB often has a much smaller search
space compared to the traditional BnB search.

The structure of an AOBB search tree is determined by its pseudo-tree.  In
order to represent the dependency between nodes, an AOBB search tree contains
two types of nodes.  The first type of nodes is called \emph{OR nodes}, which
splits the space into several parts that cover the original space by assigning
a particular rotamer to a residue.  The second type of nodes is called
\emph{AND nodes}, which decomposes the space into several smaller spaces where
the computations of total energy of residues in different branches are
independent to each other.  The root of an AOBB search tree is an OR nodes and
all the leaves are AND nodes.  For each node in an AOBB search tree, its type
is different from that of its parent. An example of an AOBB search tree is
given in \Fref{fig:aotree}.

Unlike the traditional BnB search, in which a solution is represented by a
single leaf node, in an AOBB search tree, a valid conformation is represented
by a tree, called the \emph{solution tree}.  A solution tree shares the same
root with the AOBB search tree. If an AND node is in the solution tree, all its
OR children are also in the tree.  If an OR node is in the solution tree, exact
one of its AND children is in the tree.  The tree with bold lines in
\Fref{fig:aotree} shows an example of a solution tree.  In order to compute the
best solution tree with the minimum energy when traversing the search space, we
can maintain a \emph{node value} $v(x)$ to store the total energy involving the
residues in the sub-tree rooted at $x$.  In an AOBB search tree, $v(x)$ can be
computed as follows:
\begin{equation}
  v(x) = \begin{cases*}
    0, & if $x$ is a leave node; \\
    \sum_{y \in \mathrm{child}(x)} v(y), & if $x$ is an internal AND node; \\
    \min_{y \in \mathrm{child}(x)} e(y) + v(y), & if $x$ is an internal OR node, \\
  \end{cases*}
  \label{eq:v}
\end{equation}
where $\mathrm{child}(x)$ stands for the set of children of node $x$ and $e(y)$
is the sum of the self energy of the rotamer represented by $y$ and the
pairwise energy between the rotamer represented by $y$ and other rotamers
represented by the ancestors of $y$.  Then the $v(\cdot)$ value of the root of
the whole search tree is equal to the energy of the GMEC solution.  The
corresponding best solution tree can be constructed using a similar method.

\n{
  Because AOBB uses the depth-first-search strategy, its space complexity is
  $O(n)$, where $n$ is the number of mutable residues.  The time complexity of
  AOBB in the worst case is $O(n*p^d)$, where $p$ is the number of rotamers per
  residue and $d$ is the depth of the pseudo-tree.  A more detailed explanation
  about the AOBB search with pseudocode can be found in Appendix Section A2
  \cite{zhou2015appendix}.
}


\subsubsection{Heuristic Function} \label{sec:heuristic} 
The choice of the heuristic function $h(x)$, \n{which is a lower bound of
$v(x)$}, heavily affects the performance of the AOBB algorithm.  A popular
heuristic function used with AOBB is called \emph{mini-bucket heuristic}
\cite{kask2001general}, which is computed by the \emph{mini-bucket elimination}
algorithm \cite{dechter2003mini}.  The computation of mini-bucket heuristic can
be accelerated through pre-computation, so that $h(x)$ can be computed
efficiently by looking up of pre-computed tables.  The bound given by the
mini-bucket heuristic can be further \n{tighten by Max-Product Linear
Programming \cite{globerson2008fixing} and Join Graph Linear Programming
\cite{ihler2012join}.}

\begin{figure}[htpb] 
  \centering  
  \subfigure[Bucket elimination for a pseudo-tree]{
    \begin{tikzpicture}[
        font=\scriptsize,
        node/.style={
          circle,
          draw,
        },
        label distance=-3pt,
      ]
      \node[node, label=above:{$h_A()$    }] (A) {A};
      \node[node, position=-130: 7mm from A, label=right:{$h_B(r_A)$    }] (B) {$B$};
      \node[node, position=-130: 7mm from B, label=right:{$h_C(r_A,r_B)$  }] (C) {$C$};
      \node[node, position=-130: 7mm from C, label=below:{$h_D(r_A,r_B,r_C)$}] (D) {$D$};
      \node[node, position= -50: 7mm from C, label=below:{$h_E(r_B,r_C)$  }] (E) {$E$};
      \node[node, position= -50: 7mm from A, label=below:{$h_F(r_A)$    }] (F) {$F$};
      \node[node, position= -50: 7mm from F, label=below:{$h_G(r_A,r_F)$  }] (G) {$G$};
      \path
        (A) edge (B) edge (F) edge[dashed,in=90, out=180] (D) edge[dashed, in=90, out=0] (G)
        (B) edge (C) edge[dashed, in=90, out=180] (D) edge[dashed, in=45, out=-45] (E)
        (C) edge (D) edge (E)
        (F) edge (G)
        ;
    \end{tikzpicture}
    \label{fig:bucket}
  }
  \subfigure[Mini-bucket heuristic]{
    \begin{tikzpicture}[
        font=\scriptsize,
        node/.style={
          circle,
          draw,
        },
        shade/.style={
          fill=black!10,
        },
        label distance=-3pt,
      ]
      \node[node, label=above:{$h'_A()$    }] (A) {A};
      \node[node, position=-130: 7mm from A, label=right:{$h'_B(r_A)$    }] (B) {$B$};
      \node[node, position=-130: 7mm from B, label=right:{$h'_C(r_B)$  }] (C) {$C$};
      \node[node, position=-130: 7mm from C, label=below:{$h'_D(r_B,r_C)$},shade] (D) {$D$};
      \node[node, position= 90: 13mm from D, label=above:{$h'_{D'}(r_A)$},shade, inner sep=3pt] (D') {$D'$};
      \node[node, position= -50: 7mm from C, label=below:{$h'_E(r_B,r_C)$  }] (E) {$E$};
      \node[node, position= -50: 7mm from A, label=below:{$h'_F(r_A)$    }] (F) {$F$};
      \node[node, position= -50: 7mm from F, label=below:{$h'_G(r_A,r_F)$  }] (G) {$G$};
      \path
        (A) edge (B) edge (D') edge  (F) edge[dashed, in=90, out=0] (G)
        (B) edge (C) edge[dashed, in=90, out=180] (D) edge[dashed, in=45, out=-45] (E)
        (C) edge (D) edge (E)
        (F) edge (G)
        ;
    \end{tikzpicture}
    \label{fig:minibucket}
  }
  \caption{
    An example of mini-bucket elimination.  (a) The pseudo-tree of a graph
    along with the resulting energy tables computed by the bucket elimination
    algorithm.  The dashed lines represents the non-tree edges in the original
    graph.  (b) The tree generated by the mini-bucket elimination algorithm for
    the pseudo-tree in (a), in which the original energy table $h_D(r_A, r_B,
    r_C)$ is split into two smaller tables $h'_D(r_B, r_C)$ and $h'_{D'}(r_A)$.
  }
  \label{fig:bucketall}
\end{figure}
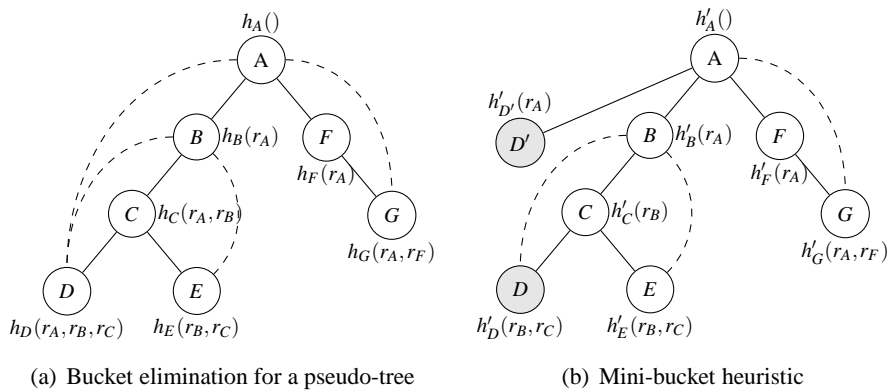 

The mini-bucket elimination is an approximation version of the \emph{bucket
elimination} algorithm \cite{dechter1998bucket},  which is an another exact
algorithm for solving the combinatorial problem with an underlying graph
structure, such as protein design, based on a pseudo-tree.  More specifically,
the bucket elimination algorithm maintains an energy table $h_x(\cdot)$ for
each tree node $x$,  which stores the exact lower bound on the sum of energy
involving the residues in the sub-tree rooted at $x$ given the rotamer
assignments of $x$'s ancestors.  For instance, $h_D(r_A, r_B, r_C)$ in
\Fref{fig:bucket} stores the exact lower bound of node $D$ given the rotamer
assignments of its ancestors $r_A$, $r_B$, and $r_C$.  These energy tables can
be computed in a bottom-up manner.  As an example, \Fref{fig:bucket} shows the
energy tables of the bucket elimination on a pseudo-tree of a residue
interaction network,  and we can compute $h_C(r_A, r_B) = \min_{r_C}(E(r_B,
r_C) + h_D(r_A, r_B, r_C) + h_E(r_B, r_C))$, where $E(r_B, r_C)$ represents the
pairwise energy term between rotamers $r_B$ and $r_C$.  The $h$ value of the
tree root, $h_A()$, in this example, is the total energy of the GMEC.  The time
complexity of bucket elimination is $O(n*\exp(w))$ \cite{dechter1998bucket},
where $n$ is the number of the nodes and $w$ is the tree width
\cite{robertson1986graph} of the graph.

\n{
  If the tree width of a graph is large, the energy tables may have high
  dimensions and thus can be too large to compute.  The mini-bucket elimination
  is proposed to address this problem.  In particular, it splits a node with a
  large energy table into multiple nodes with smaller energy tables, called
  \emph{mini-buckets}, along with the pairwise energy term represented by the
  new added edges to decrease the dimension of its original energy table.  We
  use $h'_x(\cdot)$ to represent the new energy table for each node $x$
  computed by the mini-bucket algorithm.  \Fref{fig:minibucket} gives an
  example, in which $h_D(r_A, r_B, r_C)$ is split into two smaller tables
  $h'_D(r_B, r_C) = \min_{r_D}(E(r_D, r_B) + E(r_D, r_C))$ and $h'_{D'}(r_A) =
  \min_{r_D} E(r_D, r_A)$.  Because now $D$ and $D'$ can be assigned with
  different rotamers, the new energy tables computed by the bucket elimination
  on the new graph is a lower bound of the original problem.  Therefore, we can
  use the sum of $h'_x(\cdot)$ on all mini-buckets of a node as the heuristic
  function for AOBB.
}

%

\subsection{Finding Sub-optimal Conformations} \label{sec:subopt} 
\n{
In practice, we often require the design
algorithm to output the $k$ best conformations within a given energy cutoff
$\Delta$ \cite{donald2011algorithms}.  In the BnB framework, this can be done
easily by running the BnB search $k$ times and remove the optimal conformations found in the preceding rounds
from the search space.  The task is more complicated to tackle in the AOBB
because a conformation is represented by a solution tree
rather than a tree node.  Our solution consists of two parts:
\begin{compactenum}
  \item In bounding steps, do not prune nodes in which the heuristic
    function values of the corresponding solution trees do not exceed the
    critical value by $\Delta$.
  \item Keep track of the $k$ best solution trees and their $v(\cdot)$ values
    rather than only a single solution.
\end{compactenum}
For the second part, we need to extend the procedure of computing $v(x)$,
originally described in \Fref{eq:v}.  For each node $x$, we now store the
$k$ node values.  Let $v_1(x)$ be the best node value, $v_2(x)$ be the second
one, and so on.  For each leaf node $x$, $v_1(x) = 0$ and $v_2(x) = v_3(x) =
\dots = v_k(x) = \infty$.  For each OR node $x$, we can compute $v_1(x) \le
v_2(x) \le \dots \le v_k(x)$ by merging $v_i(\cdot)$ values of $x$'s
children using a sort routine and retaining the $k$ smallest values.
}

The merge operation for AND nodes is very challenging.  For each AND node $x$,
let its children be $y_1$, $y_2$, \dots, $y_t$.  Our task is to find $k$
different sequences $(a_1, \dots, a_j, \dots, a_k)$, where $a_j = (a_{j1},
a_{j2}, \dots, a_{jt})$ and $a_{ji} \in \{1, 2, \dots, k\}$, so that $v_j(x) =
\sum_{i=1}^t v_{a_{ji}}(y_i)$ and $v_1(x) \le v_2(x) \le \dots \le v_k(x)$.  A
brute-force method for solving this problem requires $O(k^t)$ time complexity
as it needs to enumerate all possible sequences for $a_1,a_2,\dots,a_k$, which
is unacceptable because both $k$ and $t$ may be large in a real problem.

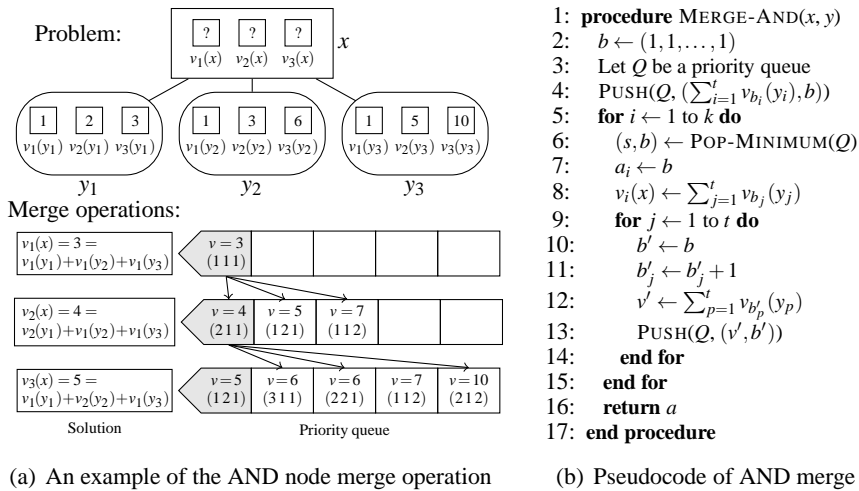
\begin{figure}[htpb] 
  \centering
  \subfigure[An example of the AND node merge operation]{
    \begin{tikzpicture}[
        font=\tiny,
        node distance=7pt,
        label distance=-2pt,
        box/.style={
          draw,
          rectangle,
          inner sep=0,
          minimum size=10pt,
        },
        bq/.style={
          draw,
          rectangle,
          align=center,
          inner sep=2pt,
          minimum height=17pt,
          minimum width=23.5pt,
        },
        head/.style={
          signal,
          signal to=west,
          fill=black!10,
        },
      ]
      \node[box,label=below:$v_2\!(x)$] (rm) {?};
      \node[box,label=below:$v_1\!(x)$] (r1) [left=of rm] {?};
      \node[box,label=below:$v_3\!(x)$] (r2) [right=of rm] {?};

      \node[box,label=below:$v_2\!(y_2)$] (bm) [below=.8cm of rm] {$3$};
      \node[box,label=below:$v_1\!(y_2)$] (b1) [left=of bm] {$1$};
      \node[box,label=below:$v_3\!(y_2)$] (b2) [right=of bm] {$6$};

      \node[box,label=below:$v_2\!(y_1)$] (am) [left=1.8cm of bm] {$2$};
      \node[box,label=below:$v_1\!(y_1)$] (a1) [left=of am] {$1$};
      \node[box,label=below:$v_3\!(y_1)$] (a2) [right=of am] {$3$};

      \node[box,label=below:$v_2\!(y_3)$] (cm) [right=1.8cm of bm] {$5$};
      \node[box,label=below:$v_1\!(y_3)$] (c1) [left=of cm] {$1$};
      \node[box,label=below:$v_3\!(y_3)$] (c2) [right=of cm] {$10$};

      \node [left=0.8cm of r1] {\footnotesize Problem:};

      \begin{pgfonlayer}{background}
        \node[
          rectangle, draw, inner sep=8pt, yshift=-4pt,
          label=right:\footnotesize$x$
        ] (rr) [fit = (rm) (r1) (r2)] {};
        \node[
          rectangle, draw, inner sep=5pt, yshift=-5pt,
          minimum height=1.1cm, label=below:\footnotesize$y_1$,
          rounded corners=14pt,
        ] (aa) [fit = (am) (a1) (a2)] {};
        \node[
          rectangle, draw, inner sep=5pt, yshift=-5pt,
          minimum height=1.1cm, label=below:\footnotesize$y_2$,
          rounded corners=14pt,
        ] (bb) [fit = (bm) (b1) (b2)] {};
        \node[
          rectangle, draw, inner sep=5pt, yshift=-5pt,
          minimum height=1.1cm, label=below:\footnotesize$y_3$,
          rounded corners=14pt,
        ] (cc) [fit = (cm) (c1) (c2)] {};
        \path[]
          (rr) edge[shorten >=-6.2pt] (aa)
          (rr) edge (bb)
          (rr) edge[shorten >=-6.2pt] (cc)
          ;
      \end{pgfonlayer}

      \node[draw, below=13mm of am, inner sep=1pt, xshift=2pt] (p1)
        {\setlength{\jot}{0pt}$\begin{aligned}
          &v_1\!(x) = 3 = \\
          {}&v_1\!(y_1) \!+\! v_1\!(y_2) \!+\!v_1\!(y_3)
        \end{aligned}$};
      \node[draw=none, above=0pt of p1] { \footnotesize Merge operations: };
      \node[draw, right=2pt of p1, bq,head] (q11)
        { $v=3$\\$(1\,1\,1)$};
      \node[draw, right=-0.4pt of q11, bq] (q12)
        {};
      \node[draw, right=-0.4pt of q12, bq] (q13)
        {};
      \node[draw, right=-0.4pt of q13, bq] (q14)
        {};
      \node[draw, right=-0.4pt of q14, bq] (q15)
        {};

      \node[draw,below=3mm of p1, inner sep=2pt] (p2)
        {\setlength{\jot}{0pt}$\begin{aligned}
          &v_2\!(x) = 4 = \\
          {}&v_2\!(y_1) \!+\! v_1\!(y_2) \!+\!v_1\!(y_3)
        \end{aligned}$};
      \node[draw, right=2pt of p2, bq, head] (q21)
        { $v=4$\\$(2\,1\,1)$ };
      \node[draw, right=-0.4pt of q21, bq] (q22)
        { $v=5$\\$(1\,2\,1)$ };
      \node[draw, right=-0.4pt of q22, bq] (q23)
        { $v=7$\\$(1\,1\,2)$ };
      \node[draw, right=-0.4pt of q23, bq] (q24)
        {};
      \node[draw, right=-0.4pt of q24, bq] (q25)
        {};

      \node[draw,below=3mm of p2, inner sep=1pt] (p3)
        {\setlength{\jot}{0pt}$\begin{aligned}
          &v_3\!(x) = 5 = \\
          {}&v_1\!(y_1) \!+\! v_2\!(y_2) \!+\!v_1\!(y_3)
        \end{aligned}$};
      \node[draw, right=2pt of p3, bq, head] (q31)
        { $v\!=\!5$\\$(1\,2\,1)$ };
      \node[draw, right=-0.4pt of q31, bq] (q32)
        { $v\!=\!6$\\$(3\,1\,1)$ };
      \node[draw, right=-0.4pt of q32, bq] (q33)
        { $v\!=\!6$\\$(2\,2\,1)$ };
      \node[draw, right=-0.4pt of q33, bq] (q34)
        { $v\!=\!7$\\$(1\,1\,2)$ };
      \node[draw, right=-0.4pt of q34, bq] (q35)
        { $v\!=\!10$\\$(2\,1\,2)$ };

      \path[->]
        (q11.south) edge (q21.north) edge (q22.north) edge (q23.north)
        (q21.south) edge (q32.north) edge (q33.north) edge (q35.north)
        ;
      \node[below=0 of p3] {Solution};
      \node[below=0 of q33] {Priority queue};
    \end{tikzpicture}
    \label{fig:merge}
  }
  \hspace{5pt}
  \subfigure[Pseudocode of AND merge]{
    \begin{minipage}[b]{.35\textwidth}
    \scriptsize
    \algrenewcommand\algorithmicindent{0.8em}
    \begin{algorithmic}[1]
      \Procedure{Merge-And}{$x$, $y$}
        \State $b \gets (1, 1, \dots, 1)$
        \State Let $Q$ be a priority queue
        \State \Call{Push}{$Q$, $(\sum_{i=1}^t v_{b_i}(y_i), b)$} \label{ln:minit}
        \For{$i \gets 1$ to $k$}
          \State $(s, b) \gets$ \Call{Pop-Minimum}{$Q$} \label{ln:pop}
          \State $a_i \gets b$
          \State $v_i(x) \gets \sum_{j=1}^t v_{b_j}(y_j)$
          \For{$j \gets 1$ to $t$} \label{ln:exp1}
            \State $b' \gets b$
            \State $b'_{j} \gets b'_{j} + 1$
            \State $v' \gets \sum_{p=1}^t v_{b'_p}(y_p)$
            \State \Call{Push}{$Q$, $(v', b')$}
          \EndFor \label{ln:exp2}
        \EndFor
        \State \Return $a$
      \EndProcedure
    \end{algorithmic}
    \label{alg:merge}
  \end{minipage}
  }
  \caption{
    The merge operation for AND nodes.  (a) An example where the upper part
    describes the problem and the lower part shows how to solve this problem
    using a priority queue.  The numbers in small squares show the
    corresponding $v(\cdot)$ values of individual tree nodes.  The shaded boxes
    show the element with the smallest value in each priority queue. (b) The
    pseudocode of the merge operation for AND nodes.
  }
  \label{fig:merge-2in1}
\end{figure}

A simple example and the pseudocode of our merge operation for an AND node are
shown in in \Fref{fig:merge-2in1}.  This algorithm uses a priority queue $Q$,
which is a data structure that supports the operations of inserting a key/value
pair (i.e., element) and extracting the element with the minimum value.  We
first define an index sequence $b=(b_1, \dots, b_t)$, where entry $b_i$
represents the index of the chosen $v(\cdot)$ value in child $y_i$.  Initially,
$b = (1, 1, \dots, 1)$ is pushed to $Q$.  In this problem, the value of an
element is the sum of $v(\cdot)$ values of the AND nodes' children computed
using the index sequence $b$ as the key (\Fref{ln:minit}).  The initial index
sequence $b=(1, 1, \dots, 1)$ corresponds to the first sequence $a_1$ because
we choose the best $v$ value for each child and thus we can get the best
$v(\cdot)$ value for their parent.  Each time we extract the element with the
minimum value from $Q$ as the next best sequence (\Fref{ln:pop}).  Then we push
all the successors of the extracted sequence, computed by increasing only one
index for each element in the sequence, into the priority queue (Lines
\ref{ln:exp1} to \ref{ln:exp2}).  We repeat these steps until all the $a_i$
values are generated.  The time complexity of this process is $O(kt\log(kt))$.
\n{The proof of the correctness about our merge algorithm is provided in
Appendix Section A3. \cite{zhou2015appendix}}

\section{Results} \label{sec:experiment} 
We conducted two computational experiments to evaluate the performance of our
new AOBB-based protein design algorithm.  In the first experiment, we compared
our new AOBB-based algorithm with the traditional A*-based algorithm in a core
redesign problem.  To make a fair comparison, in this test we did not not make
any approximation in the energy matrix (i.e., the residue interaction network
is fully connected) because the A*-based algorithm cannot benefit much from
such approximation.  In the second computational experiment, we performed the
full protein design to \n{examine the performance} of our algorithm on a larger
residue interaction network.

Our AOBB-based protein design algorithm was implemented based on the protein
design package OSPREY \cite{keedy2013osprey} and the UAI branch of the AOBB
search framework \texttt{daoopt} \cite{otten2012anytime,otten2012wining}.  For
comparison, we used the DEE/A* solver provided by the OSPREY package.  In
addition, we included the sequential A* solver with the improved computation of
heuristic functions \cite{zhou2014efficient}.  We used an Intel Xeon E5-1620
3.6GHz CPU in all evaluation tests.
%

\subsection{Core Redesign} 
\n{Core redesign can replace the amino acids in
the core of a wild-type protein to increase its thermostability
\cite{korkegian2005computational}}.  In this experiment, we tested all the 23
protein core redesign cases that failed to be solved in using the expanded
rotamer library with the rigid DEE/A* in 4G memory \n{from} \cite{gainza2012protein}.
In addition, we picked another 5 design problems from \cite{gainza2012protein}
that were solvable \n{within} the given memory using the traditional DEE/A* algorithm.
To make a fair comparison between A* and AOBB search algorithm, we did not
remove any edge from the fully connected residue interaction network during the
AOBB search in this test.

\begin{table}[htpb]
  \scriptsize
  \centering
  \begin{tabular}{crrrrrr}
    \hline
    PDB & Space size & \;\; \# of A* states & \;\; \# of AOBB states &
    \;\; \n{OSPREY time} & \;\; \n{cOSPREY time} & \;\; AOBB time \\
    \hline
    \texttt{1TUK} & 1.73e+19 & OOM & 188,042 & OOM & OOM & 723 \\
    \texttt{1ZZK} & 3.44e+15 & OOM & 255 & OOM & OOM & $<1$ \\
    \texttt{2BWF} & 5.54e+22 & OOM & 517,258,245 & OOM & OOM & 1,467,951 \\
    \texttt{3FIL} & 2.62e+21 & OOM & 3 & OOM & OOM & $<1$ \\
    \texttt{2RH2} & 1.29e+22 & OOM & OOT & OOM & OOM & OOT \\
    \hline
    \texttt{1IQZ} & 7.11e+17 & 18,337,117 & 90,195 & 1,824,235 & 40,217 & 117 \\
    \texttt{2COV} & 1.14e+10 & 43,306 & 3 & 317 & 21 & 1 \\
    \texttt{3FGV} & 6.44e+12 & 3,073,965 & 3 & 59,589 & 5,091 & $<1$ \\
    \texttt{3DNJ} & 5.11e+12 & 569,597 & 4,984 & 7,469 & 570 & 3 \\
    \texttt{2FHZ} & 1.83e+18 & 14,732,913 & 3,972 & 3,475,716 & 70,783 & 13 \\
    \hline
  \end{tabular}
  \caption{The comparison between A*-based and AOBB-based algorithms on core redesign}
  \label{tab:redesign}
  \vspace{-10pt}
\end{table}

\Fref{tab:redesign} summarizes the comparison results between A*-based and our
AOBB-based algorithms, in which ``OOM''and ``OOT'' represents ``out of memory''
and ``out of time'', respectively.  The full comparison results can be found in
Appendix Table A1 \cite{zhou2015appendix}.   The memory was limited to 4G, which was the same as
that in \cite{gainza2012protein}, and the running time was limited to 8 hours.
The first five rows show the five cases (among 23 cases) in
\cite{gainza2012protein} which were formerly unsolvable by the original A*
algorithm. The column labeled with ``Space size'' shows the size of the
conformational space after DEE pruning.  \n{The columns labeled with ``OSPREY
time'' and ``cOSPREY time'' show running time of the A* solvers from
OSPREY and \cite{zhou2014efficient}, respectively}.  The running time was
measured in millisecond and did not include the initialization steps of each
algorithm.  The initialization time of AOBB was relatively stable for all cases
and typically took 90s \n{to} compute the mini-bucket heuristic tables
and an initial bound for AOBB search.

As shown in \Fref{tab:redesign} and Table A1, the AOBB algorithm can
successfully find the GMEC solutions for 21 out of the 23 problems from
\cite{gainza2012protein}'s data, which were formerly unsolvable by the original
A* algorithm in 4G memory.  Also, we find that the number of states expanded in
the AOBB search was much less than that in the traditional A* search.
Accordingly, for those cases solvable by both algorithms, the AOBB search
consumed less time than the traditional A* search.  \n{Probably this
improvement was due to the fact that the mini-bucket heuristic with MPLP and
JGLP is tighter than the heuristic function used in OSPREY.}


\subsection{Full Protein Design} 
In the second computational experiment, we ran the full protein design to
evaluate the performance of our AOBB-based protein design algorithm.  In the
full protein design problem, all residues of a protein are mutable, which leads
to a much larger conformational space.  For each residue, we picked 1-4 the
most similar amino acids, according to the BLOSUM62 matrix, as the mutation
candidates.  For each pair of residues $A$ and $B$, we added an edge $(A, B)$
to the residue interaction network if and only if for all rotamer assignments
$r_A$ and $r_B$, $(\max_{r_A, r_B}E(r_A, r_B) - \min_{r_A, r_B}E(r_A, r_B)) >
\lambda$, where threshold parameter $\lambda$ was used to trade the precision of
the energy with the easiness of the problem.  We used $\lambda=0.04$ for all
the test cases.

\begin{table}[htbp]
  \centering
  \footnotesize
  \begin{tabular}{crrrrrr}
    \hline
    PDB & \;\; Space size & \;\;\# of residues & \;\;\# of edges & \;\; Tree depth & \;\; \# of AOBB states & \;\; AOBB time \\
    \hline
    \texttt{1I27} & 6.69e+45 & 69 & 968 & 40 & 3,149 & 11 \\
    \texttt{1M1Q} & 2.33e+19 & 71 & 390 & 17 & 3 & $<1$ \\
    \texttt{1T8K} & 2.83e+43 & 75 & 1031 & 42 & 3 & $<1$ \\
    \texttt{1XMK} & 2.66e+48 & 74 & 1108 & 40 & 864 & 2 \\
    \texttt{3G36} & 4.28e+20 & 47 & 396 & 22 & 159 & $<1$ \\
    \texttt{3JTZ} & 1.96e+45 & 71 & 961 & 44 & 4,354,110 & 17,965 \\
    \hline
  \end{tabular}
  \caption{The test results on the full protein design problem}  \label{tab:full}
  \vspace{-10pt}
\end{table}
%

\n{\Fref{tab:full} shows the test results of this computational experiment.  The
running time was measured in millisecond.  Here we did not list the results of
traditional A*-based algorithm because we found that A*-based algorithms were
unable to find the GMEC solutions for all these test cases within 4G memory.
The AOBB-based search algorithm can found the GMEC solutions for all the test
cases.  This demonstrates the power of \n{the AOBB search algorithm with the
state-of-the-art heuristic function, which can effectively address full protein
design problems}.
}

\section{Conclusion and Future Work} \label{sec:conclusion}  
In this paper, we developed a new protein design algorithm based on the new
branch-and-bound search technique (i.e., AOBB) to find the global minimum
energy conformation, which speeds up the search process by several orders of
magnitude than the traditional provable algorithms. The AOBB-based algorithm
accelerates the search process based on an advanced heuristic function and
fully exploits the topology of the residue interaction network while it only
has linear memory consumption.  The algorithm can also output suboptimal
solutions by employing an elegant modification of the original search
algorithm.

Currently, our algorithm is only implemented on a single machine. It is
possible to further accelerate the design process by parallelizing AOBB search
on a GPU processor or a CPU cluster on a supercomputer, which will enable us to
deal with the protein design problems with larger conformational space.

\section*{Acknowledgement} 
We thanks Dr. Lars Otten and Prof. Alex Ihler for their support in providing
their code of the \texttt{daoopt} AOBB solver.

\noindent\textbf{Funding:} This work was supported in part by the National Basic
Research Program of China Grant 2011CBA00300, 2011CBA00301, the National
Natural Science Foundation of China Grant 61033001, 61361136003 and 61472205,
and China's Youth 1000-Talent Program.

\begin{spacing}{0.95}
\bibliography{ref.bib}
\end{spacing}

\end{document}